\documentclass{article}

\usepackage{times}
\usepackage{graphicx} 
\usepackage{subfigure} 

\usepackage{natbib}

\usepackage{algorithm}
\usepackage{algorithmic}

\usepackage{hyperref}

\usepackage[accepted]{icml2014}

\icmltitlerunning{A Kernel Independence Test for Random Processes}

\usepackage{amssymb,amsmath}
\usepackage{amsthm}
\usepackage[usenames,dvipsnames]{xcolor}
\usepackage{enumitem}

\newtheorem{lemma}{Lemma}
\newtheorem{Theorem}{Theorem}

\newtheorem{statement}{Statement}
\newtheorem{corollary}{Corollary}

\newcommand{\ev}{\mathcal{E}}

\begin{document} 


\twocolumn[
\icmltitle{A Kernel Independence Test for Random Processes}

\icmlauthor{Kacper Chwialkowski}{kacper.chwialkowski@gmail.com}
\icmladdress{University College London, Computer Science Department}
\icmlauthor{Arthur Gretton}{arthur.gretton@gmail.com}
\icmladdress{University College London, Gatsby Computational Neuroscience Unit}

\icmlkeywords{HSIC, Random processes, time series, independence}

\vskip 0.3in
]

\begin{abstract} 
A non-parametric approach to the problem of testing the independence of two random processes is developed.  The test statistic is the Hilbert-Schmidt Independence Criterion (HSIC), which was used previously in testing independence for i.i.d. pairs of variables. The asymptotic behaviour of HSIC is established when computed from samples drawn from random processes. It is shown that earlier bootstrap procedures which worked in the i.i.d. case will fail for random processes, and an alternative consistent estimate of the p-values is proposed. Tests on artificial data and real-world forex data indicate that the new test procedure discovers dependence which is missed by linear approaches, while the earlier bootstrap procedure returns an elevated number of false positives. The code is available online: \url{https://github.com/kacperChwialkowski/HSIC}.
\end{abstract} 

\vspace{-3mm}
\section{Introduction}

Measures of statistical dependence between pairs of random variables $(X,Y)$ are well established, and have been applied in a wide variety of areas, including  fitting causal networks \cite{Pearl01},  discovering features which have significant dependence on a label set \cite{SonSmoGreBedetal12}, and independent component analysis \cite{HyvKarOja01}. Where pairs of observations are independent and identically distributed, a number of non-parametric tests of independence have been developed \cite{Feuerverger93,gretton_kernel_2008,SzeRiz09,GreGyo10}, which determine whether the dependence measure value is  statistically significant. These non-parametric tests are consistent against any fixed alternative - they make no assumptions as to the nature of the dependence. 

For many data analysis tasks, however, the observations being tested are drawn from a time series: each observation is dependent on its past values. Examples include audio signals, financial data, and brain activity.  Given two such random processes, we propose a hypothesis test of instantaneous dependence, of whether the two signals are dependent at a particular time $t$. Our test satisfies two important properties: it is consistent against any fixed alternatives, and it is non-parametric - we do not assume the dependence takes a particular form (such as linear correlation), nor do we require  parametric models of the time series. We further avoid making use of a density estimate as an intermediate step, so as to avoid the assumption that the distributions have densities (for instance, when dealing with text or other structured data).

We use as our test statistic the Hilbert-Schmidt Independence Criterion (HSIC) \citep{gretton_measuring_2005,gretton_kernel_2008}, which can be interpreted as the distance between embeddings  of the joint distribution and the product of the marginals in a reproducing kernel Hilbert space (RKHS) \citep[Section 7]{gretton_kernel_2012}. When characteristic RKHSs are used, the HSIC is zero iff the variables are independent \citep{SriGreFukLanetal10}. Under the null hypothesis of independence, $P_{XY}=P_X P_Y$, the minimum variance estimate of HSIC is a degenerate U-statistic. The distribution of the empirical HSIC under the null is an infinite sum of independent $\chi^2$ variables \cite{gretton_kernel_2008}, which follows directly from e.g. \citep[Ch. 5]{serfling_approximation_2002}. In practice, given a sample $(x_i,y_i)_{i=1}^n$ of pairs of variables drawn from $P_{XY}$,
 the null distribution is approximated by a bootstrap procedure, where a histogram is obtained by computing the test statistic on many different permutations $\{x_i,y_{\pi(i)}\}_{i=1}^n$,  to decouple $X$ and $Y$.

In the case where the samples $Z_t=(X_t,Y_t)$ are drawn from a random process, the analysis of the asymptotic behaviour of HSIC requires substantially more effort than in the i.i.d. case. As our main contribution, we obtain both the null and alternative  distributions of HSIC for random processes, where the null distribution is defined as $X_t$ being independent of $Y_t$ at time $t$. Such a test may be used for rejecting causal effects (i.e., whether one signal is not dependent on the values of another signal at a particular delay)  or instant coupling (see our first experiment in Section \ref{sec:forex}).\footnote{We distinguish our case from the problem of ensuring time series are independent simultaneously across all time lags, e.g the null will hold even if $X_t = Y_{t-1}$ where $ Y_{t}$ is white noise.} The null distribution is again an infinite weighted sum of $\chi^2$ variables, however these are now correlated, rather than independent. Under the alternative hypothesis, the statistic has an asymptotically normal distribution. 

For the test to be used in practice, we require an empirical estimate of the null distribution, which gives the correct test threshold when $Z_t=(X_t,Y_t)$ is a random process. Evidently, the bootstrap procedure used in the i.i.d. case is  incorrect, as the temporal dependence structure within the $Y_t$ will be removed. This turns out to cause severe problems in practice, since the permutation procedure will give an increasing rate of false positives as the temporal dependence of the $Y_t$ increases  (i.e., dependence will be detected between $X_t$ and $Y_t$, even though none exists, this is also known as a Type I error). Instead, our null estimate is obtained by making shifts of one signal relative to the other, so as to retain the dependence structure within each signal. Consequently, we are able to keep the Type I error  at the designed level $\alpha = 0.05$. In our experiments, we address three examples: one artificial case consisting of two signals which are dependent but have no  correlation, and two real-world examples on forex data.  HSIC for random processes reveals dependencies that classical approaches fail to detect. Moreover, our new approach  gives the correct Type I error rate, whereas a bootstrap-based approach designed for i.i.d. signals returns too many false positives.

\vspace{-2mm}
\paragraph{Related work}

Prior work on testing independence in time series may be categorized in two branches: testing serial dependence within a single time series, and testing dependence between one time series and another.  The case of serial dependence turns out to be relatively straightforward, as under the null hypothesis, the samples become independent: thus, the analysis reduces to the i.i.d. case. \citet{pinkse_consistent_1998,diks_nonparametric_2005} provide a quadratic forms function-based serial dependence test which employs the same statistic as HSIC. Due to the simple form of the null hypothesis, the analysis of \citep[Ch. 5]{serfling_approximation_2002} applies.  Further work in the context of the serial dependency testing includes simple approaches based on rank statistics e.g. Spearman's correlation or Kendall's tau, correlation integrals e.g. \cite{broock_test_1996}; criteria based on integrated squared distance between densities e.g \cite{rosenblatt_nonparametric_1992}; KL-divergence based criteria  e.g. \cite{robinson_consistent_1991,hong_asymptotic_2004}; and generalizations of KL-divergence to so called $q$-class entropies e.g. \cite{clive_w._j._granger_dependence_2004,racine_versatile_2007}.

In most of the  tests of independence of two time series, specific conditions have been enforced, e.g that  processes follow a moving average specification or the dependence is linear.
 Prior work in the  context of dependency tests of two time series includes cross covariance based tests e.g. \cite{haugh_checking_1976,hong_testing_1996,shao_generalized_2009}; and a Generalized Association Measure based criterion \cite{fadlallah_association_2012}. Some work has  been undertaken in the non-parametric case, however.
 A non-parametric measure of independence for time series, based on the Hilbert-Schmidt Independence Criterion, was proposed by \citet{smola_kernel_2008}. While this work established the convergence in probability of the statistic to its population value, no asymptotic distributions were obtained, and the statistic was not used in hypothesis testing.
To our knowledge, the only non-parametric independence test for pairs of time series is due to \citet{besserve_statistical_2013}, which addresses the harder problem of  testing independence across all time lags simultaneously. \footnote{ Let $X_t$ follow a MA(2) model and put $Y_t = X_{t-20}$. This is a case addressed by \citet{besserve_statistical_2013}, who will reject their null hypothesis, whereas our null is accepted} The procedure is to compute the Hilbert-Schmidt norm of a cross-spectral density operator (the Fourier transform of the covariance operator at each time lag). The resulting statistic is a function of frequency, and must be zero at all frequencies for independence, so a correction for multiple hypothesis testing is required. It is not clear how the asymptotic analysis used in the present work would apply to this statistic, and this remains an interesting topic of future study.


The remaining material is organized as follows. In Section \ref{sec:intro} we  provide a brief introduction to random processes and various mixing conditions, and an expression for our independence statistic, HSIC. In Section \ref{sec:math}, we
characterize the asymptotic behaviour of  HSIC for random variables with temporal dependence, under the null and alternative hypotheses, and establish the test consistency.  We propose an empirical procedure for constructing a statistical test, and demonstrate that the earlier bootstrap approach will not work for our case. Section \ref{sec:experiments} provides experiments on synthetic and real data.



\vspace{-2mm}
\section{Background}
\label{sec:intro}
In this section we introduce necessary definitions referring to random processes. 
We then go on to define a V-statistic estimate of the Hilbert-Schmidt Independence Criterion, which applies in the i.i.d. case.

\vspace{-2mm}
\paragraph{Random process.}
First, we introduce the probabilistic tools needed for pairs of time series. Let $(Z_t,\mathcal{F}_t)_{t \in \mathbb{N}}$  be a strictly stationary sequence of random variables defined on a probability space $\Omega$ with a probability measure $P$ and natural filtration $\mathcal{F}_t$. Assume that $Z_t$ denotes a pair of random variables i.e. $Z_t = (X_t,Y_t)$, where $X_t$ is defined on $\mathcal{X}$, and $Y_t$ on $\mathcal{Y}$. Each $Z_t$ takes values in a measurable Polish space $(\mathbf{Z},\mathcal{B}(\mathbf{Z}),P_{\mathbf{Z}})$. The space $\mathbf{Z}$ is a Cartesian product of two Polish spaces $\mathbf{X}$ and $\mathbf{Y}$, endowed with a natural Borel sigma field and a probability measure. 
 
We introduce a sequence of independent copies of $Z_0$, i.e., $(Z_t^*)_{t \in \mathbb{N}}$. Since $Z_t$ is stationary, $Z_t^*$ retains the dependence between random variables $X_t$ and $Y_t$, but breaks the temporal dependence.  

Next, we formalize a concept of memory of a process. A process is called absolutely regular ($\beta$-mixing) if $\beta(m) \rightarrow 0$, where 
\begin{equation*}
\beta(m) = \frac 1 2 \sup_n \sup \sum_{i=1}^{I} \sum_{j=1}^{J}  |P(A_i \cap B_j) - P(A_i)P(B_j) |.
\end{equation*}
The second supremum in the $\beta(m)$ definition is taken over all pairs of finite partitions $\{A_1,\cdots,A_I\}$  and $\{B_1,\cdots,B_J\}$ of the sample space such that $A_i \in \mathcal{A}_{1}^{n}$ and $B_j \in \mathcal{A}_{n+m}^{\infty}$, and $\mathcal{A}_{b}^{c}$ is a sigma field spanned by a subsequence, $\mathcal{A}_{b}^{c} = \sigma(Z_b,Z_{b+1}, ..., Z_{c})$. 
A process is called uniform mixing ($\phi$-mixing) if $\phi(m) \rightarrow 0$, where
\begin{equation*}
\phi(m) = \sup_n  \sup_{A \in \mathcal{A}_{1}^{n} } \sup_{B \in \mathcal{A}_{n+m}^{\infty}}  |P(B|A) - P(B)|.
\end{equation*}
Uniform mixing implies absolute regularity, i.e. $\beta(m) \leq \phi(m)$ \cite{bradley_basic_2005}.
Under technical assumptions, Autoregressive Moving Average processes --- or more generally Markov Chains --- are absolutely regular or uniformly mixing \cite{doukhan_markov_1994}.

\vspace{-2mm}
\paragraph{Hilbert-Schmidt Independence Criterion}
\label{sec:definitions}
Let  $k$, $l$ be positive definite  kernels associated with respective reproducing kernel Hilbert spaces $\mathcal{H}_{\mathcal{X}}$ on $\mathcal{X}$, and $\mathcal{H}_{\mathcal{Y}}$ on $\mathcal{Y}$. We assume that  $k$ and $l$ are bounded and continuous. We associate to the random variable $X$ a mean embedding $\mu_X(x):=\mathcal{E}_Xk(X,x)$, such that $\forall f\in \mathcal{H}_{\mathcal{X}}$, $\langle f,\mu_X\rangle_{\mathcal{H}_{\mathcal{X}}} = \mathcal{E}_X(f(X))$ \cite{BerTho04,SmoGreSonSch07}. We assume $k$, $l$ are characteristic kernels, meaning the mappings $\mu_X$ and $\mu_Y(y):=\mathcal{E}_Yl(Y,y)$ are injective embeddings of the probability measures to the corresponding RKHSs; i.e., distributions have unique embeddings \citep{FukGreSunSch08,SriGreFukLanetal10}.

We next recall a measure of statistical dependence, the Hilbert-Schmidt Independence Criterion (HSIC), which can be expressed in terms of expectations of  RKHS kernels \citep{gretton_measuring_2005,gretton_kernel_2008}.
Denote a group of permutations over 4 elements  by $S_4$, with $\pi$ one of its elements, i.e., a permutation of four elements. We define 
\begin{equation*}
\begin{split}
h(&z_1,z_2,z_3,z_4) = \frac{1}{4!} \sum_{\pi \in S_4}  k(x_{\pi(1)},x_{\pi(2)}) [  l(y_{\pi(1)},y_{\pi(2)}) + \\
 &+  l(y_{\pi(3)},y_{\pi(4)}) - 2  l(y_{\pi(2)},y_{\pi(3)})] . \\
\end{split}  
\end{equation*}

\begin{lemma}
\label{lem:basisRelation}
Let $\gamma$ be an expected value of the function $h$, $\gamma = \ev h(Z_1^*,Z_2^*,Z_3^*,Z_4^*)$. This expectation corresponds to HSIC, computed using a function symmetric in its arguments. For $k$ and $l$ characteristic, continuous, translation invariant, and vanishing at infinity, $\gamma$ is equal to zero if and only if the null hypothesis holds (see \citep[Lemma 3.8]{Lyons13}, applying \citep[Proposition 2]{SriFukLan11}, and the note at the end of Section \ref{sec:proofs}). 
\end{lemma}
The value of  $\gamma$ corresponds to a distance between embeddings of $(X_1^*,Y_2^*)$ and $(X_1^*,Y_1^*)$ to an RKHS with the product kernel $\kappa=k\cdot l$ \citep[Section 7]{gretton_kernel_2012}. A biased empirical estimate of the Hilbert-Schmidt Independence Criterion can be expressed as a $V$-statistic (the unbiased estimate is a U-statistic, however the difference will be accounted for when constructing a hypothesis test, through an appropriate null distribution).

\vspace{-1mm}
\paragraph{$V$ statistics.} 
A $V$-statistic of a $k$-argument, symmetric function $f$ is written
\begin{equation}
\label{def:Vstat}
V(f,Z) = \frac{1}{n^k} \sum_{1 \leq i_1, \cdots, i_k \leq n} f(Z_{i_1},...,Z_{i_k}).
\end{equation}
\citet{gretton_measuring_2005} show that the biased estimator of $\gamma$ is $V(h,Z)$. The asymptotic behaviour of this statistic depends on the degeneracy of the function that defines it. We say that a $k$-argument, symmetric function $f$ is $j$-degenerate ($j<k$) if for each $z_1,\cdots,z_j \in \mathbb{Z}$,
\begin{equation*}
\ev f(z_1,\cdots , z_j , Z_{j+1}^*,\cdots ,Z_k^*) = 0.
\end{equation*}
If $j=k-1$ we say that the function is canonical. We refer to a normalized $V$ statistic as a $V$-statistic multiplied by the sample size, $n \cdot V$. 

\vspace{-2mm}
\section{HSIC for random processes}
\label{sec:math}
In this section we construct the  Hilbert-Schmidt Independence Criterion for random processes, and define its asymptotic behaviour. We then introduce an independence testing procedure  for time series.

We introduce two hypotheses: the null hypothesis  $\mathbf{H_0}$ that  $X_t$ and $Y_t$ are independent,  and the alternative hypothesis $\mathbf{H_1}$ that they are dependent. 
To build a statistical test based on  $n \cdot V(h,Z)$ we need two main results. First, if null hypothesis holds, we show $n \cdot V(h,Z)$ converges to a random variable. Second, if the null hypothesis does not hold, the $n \cdot V(h,Z)$ estimator diverges to infinity. Following these results, the Type I error (the probability of mistakenly rejecting the null hypothesis) will stabilize at the design parameter $\alpha$, and the Type II error (the probability of mistakenly accepting the null hypothesis when the variables are dependent) will drop to zero, as the sample size increases.


We begin by introducing an auxiliary kernel function  $s$, and characterize the normalized $V$-statistic distribution of $s$ using a CLT introduced by  \cite{i._s._borisov_orthogonal_2009}. We then show that the normalized $V$-statistic associated with the function $s$ has the same asymptotic distribution as the $n \cdot V(h,Z)$ distribution.  

Let $s$ be an auxiliary function
$
 s(z_1,z_2) =  \tilde{k}(x_1,x_2) \tilde{l}(y_1,y_2), 
$
where 
\begin{equation*}
\begin{split}
\tilde{k}(x_1,x_2) =& k(x_1,x_2) - \ev k(x_1,X_2) \\ 
                   &-\ev k(X_1^*,x_2) + \ev k(X_1^*,X_2^*), \\
\end{split}
\end{equation*}
and $\tilde{l}$ is defined similarly.

Both $\tilde{k}$ and $\tilde{l}$ are kernels, meaning that they are dot products between features centred in their respective RKHSs \citep{BerTho04}.  Therefore $s=\tilde{k}\cdot \tilde{l}$ defines a kernel on a product space of pairs $Z_t$. Using Mercer's Theorem we obtain an expansion for $s$. 
\begin{statement}
\label{stmt:h2}
 By \citet{steinwart_mercers_2012} Corollary 3.5,  the bounded, continuous kernel $s$  has a representation\footnote{A bounded kernel is compactly embedded into $L^2(\mathbf{Z} , \mathcal{B}(\mathbf{Z}),P_{\mathbf{Z}})$ \cite{steinwart_mercers_2012}.}
\begin{equation}
\label{eq:kernelMercer}
s(z_a,z_b) = \sum_{i=1}^{\infty} \lambda_i  e_i(z_a)e_i(z_b),
\end{equation}
where $(e_i)_{i \in \mathbb{N}^+}$ denotes an orthonormal basis  of $L^2(\mathbf{Z} , \mathcal{B}(\mathbf{Z}),P_{\mathbf{Z}})$. The series $(\sum_{i=1}^{N}$ $\lambda_i$ $e_i(z_a)e_i(z_b))$  converges absolutely and uniformly. $e_i$ are eigenfunctions of $s$ and $\lambda_i$ are eigenvalues of $s $.
\end{statement}


We will henceforth assume that for every collection of pairwise distinct subscripts $(t_1, t_2)$, the distribution of $(Z_{t_1},Z_{t_2})$ is absolutely continuous with respect to the $(Z_{t_1}^*,Z_{t_2}^*)$ distribution. This assumption prevents the occurrence of degenerate cases, such that all $Z_t$ being the same. The following results are proved in Section \ref{sec:mainProofs}.

\begin{lemma}
\label{lem:getReady}
Let the process $Z_t$ have a mixing coefficient smaller than $m^{-3}$ $(\beta(m),\phi(m) \leq m^{-3})$ and satisfy either of the following conditions:
\begin{description}
 \item[A] $Z_t$ is $\phi$-mixing.
 \item[B] $Z_t$ is $\beta$-mixing. For some $\epsilon > 0$ and for an even number $c \ge 2$, the following holds
  \begin{enumerate}[label*=\arabic*.]
    \item $\sup_i \ev |e_i(X_1)|^{2+\epsilon} \leq \infty$, where $e_i$ is the basis introduced in the Statement \ref{stmt:h2} and $|\cdot|$ denotes an absolute value.
    \item $\sum_{m=1}^{\infty} \beta^{\epsilon/(2+\epsilon)}(m) < \infty$. 
  \end{enumerate}
\end{description}
If the null hypothesis holds, then  $s$ is a canonical function and a kernel. What is more,
\begin{equation*}
  \lim_{n \to \infty} n \cdot V(s,Z) \stackrel{D}{=} \sum_{j}^{\infty} \lambda_{j} \tau_{j}^2,
\end{equation*}
where  $\tau_j$ is a centred Gaussian sequence with the covariance matrix
\begin{equation*}
\begin{split}
 \ev \tau_a \tau_b &= \ev e_a(Z_1) e_b(Z_1) + \\ 
 &+\sum_{j=1}^{\infty} \left[ \ev e_a(Z_1)e_b(Z_{j+1}) + \ev e_b(Z_1)e_a(Z_{j+1}) \right].
 \end{split}
\end{equation*}
\end{lemma}
We now characterize the asymptotics of $V(h,Z)$.
\begin{Theorem}
\label{th:main}
Under assumptions of Lemma \ref{lem:getReady}, if $\mathbf{H_0}$ holds, then the asymptotic distribution of the empirical HSIC (with scaling $n$) is the same as that of  $n \cdot V(s,Z)$,
\begin{equation*}
  \lim_{n \to \infty} n \cdot V(h,Z) \stackrel{D}{=} \lim_{n \to \infty} n \cdot V(s,Z).
\end{equation*}
\end{Theorem}  

\begin{Theorem}
\label{th:main2}
Under assumptions of the Lemma \ref{lem:getReady}, if  $\mathbf{H_1}$ holds, then $ \gamma>0$ and $ \sqrt n ( V(h,Z) - \gamma)$ has asymptotically normal distribution with mean zero and finite variance.  
\end{Theorem}  
Consequently, if the null hypothesis does not hold then $P(n \cdot V(h,Z) >C ) = P(V(h,Z) > \frac C n ) \to 1$ for any fixed $C$. Finally, we show that the $\gamma$ estimator is easy to compute. According to \citet[equation 4]{gretton_kernel_2008}, 
$
V(h,Z) = n^{-2} tr HKHL, 
$
where $K_{ab} = k(X_a,X_b)$, $L_{ab} = l(Y_a,Y_b)$ ,$H_{ij} = \delta_{ij} -  n^{-1}$ and $n$ is a sample size. 

\vspace{-2mm}
\paragraph{Testing procedure} 
%

We begin by showing that the $H_0$ distribution of the $\gamma$ estimator obtained via the bootstrap approach of \cite{diks_nonparametric_2005,gretton_kernel_2008} gives an incorrect p-value estimate when used with independent random processes. In fact, the null hypothesis obtained by permutation corresponds to the processes being {\em both} i.i.d. {\em and} independent from each other. Recall the covariance structure of the $\gamma$ estimator from Theorem \ref{th:main},
\begin{equation}
\begin{split}
\label{eq:sumCov}
 \ev \tau_a \tau_b &= \ev e_a(Z_1) e_b(Z_1) + \\ 
 &+\sum_{j=1}^{\infty} \left[ \ev e_a(Z_1)e_b(Z_{j+1}) + \ev e_b(Z_1)e_a(Z_{j+1}) \right].
 \end{split}
\end{equation}
We can represent $e_a$ and $e_b$ as $e_a(z) = e^X_u(x)e^Y_o(y)$, $e_b(z) = e^X_i(x)e^Y_p(y)$, as a decomposition  of the    $\mathbf Z$ basis into bases of $\mathbf X$ and $\mathbf Y$, respectively. Consider a partial sum $T_n$ of series from the above equation \eqref{eq:sumCov}, with $X_t$ replaced with its permutation  $X_{\pi(t)}$,    
\begin{equation}
 \begin{split}
  &T_n  =\sum_{j=1}^{n}  \ev e^X_u(X_{\pi(1)})e^X_i(X_{\pi({j+1})}) \ev e^Y_o(Y_1)e^Y_p(Y_{j+1}).
 \end{split}
\end{equation}
Using covariance inequalities from \citep[Section 1.2.2]{doukhan_markov_1994}  we conclude that $\ev e^Y_o(Y_1)e^Y_p(Y_{j+1})  = O(\Lambda(j)^{\frac 1 2 })$ and $\ev e^X_u(X_{\pi(1)})e^X_i(X_{\pi({j+1})}) = O(\Lambda(|\pi(j) - \pi(1)|)^{\frac 1 2})$ where $\Lambda$ is an appropriate mixing coefficient ($\beta$ or $\phi$). Recall that $0 < \Lambda(j) < C j^{-3}$.  

We can therefore reduce the problem to the convergence of a random variable 
\begin{equation}
\label{eq:sumTemporal}
S_n = \sum_{j=1}^{n} \Lambda(j)^{\frac 1 2} \Lambda(| \pi(j) - \pi(1)|)^{\frac 1 2} , 
\end{equation}
where $\pi$ is a random permutation drawn from the uniform distribution over the set of $n$-element permutations. In the supplementary material we show that this sum converges in probability to zero.


Since $S_n > T_n >0$, then $T_n$ converges to zero in probability,  and consequently the covariance matrix entry $ \ev \tau_a \tau_b$ converges to unity for $a = b$, and to zero otherwise. Indeed, the expected value  $\ev e_a((X_{\pi(1)},Y_1))e_b((X_{\pi(1)},Y_1)) = 0$ if $a \neq b$ and is equal to one otherwise. Note that this is the covariance matrix described by \citet{gretton_kernel_2008}.      

A correct approach to approximating the asymptotic null distribution of $n \cdot V(h,Z)$ under $\mathbf{H}_0$ is by {\em shifting}  of one time series relative to the other.  Define the shifted process $S^c_t = Y_{t+c \text{ mod } n}$ for an integer $c$, $0 \leq c \le n$ and $0 \leq t \le n$. If we let $c$ vary over $0 \leq A \leq B \le n$ for $A$ such that the dependence between $Y_{t+A}$ and $X_t$ is negligible, then we can approximate the null distribution with an empirical distribution calculated on points $(V(h,Z^k))_{ A \leq k \leq  B}$, where  $Z^k_t = (X_t,S^k_t)$. This is due to the fact that the shifted process $S^c_t$ retains most of the dependence, since it does not scramble the time index.\footnote{As a illustration, consider $W_t= Y_{t-10}$. If $Y_t$ is stationary then the dependence structure of $(W_{t_1},W_{t_2})$ and $(Y_{t_1}, Y_{t_2})$ is the same. If we set $W_t =$ $Y_{\pi(t)}$ this property does not hold.} We call this method Shift HSIC. In the supplementary material we show that Shift HSIC samples from the correct null distribution.

\vspace{-2mm}
\section{Experiments}

\label{sec:experiments}
In the experiments we compare Shift HSIC with the Bootstrap HSIC of \citet{gretton_kernel_2008}. We investigate three cases:  an artificial dataset, where two time series are coupled non-linearly; and two forex datasets, where in one case we seek residual dependence after one time series has been used to linearly predict another, and in the other case, we reveal strong dependencies between signals that are not seen via linear correlation.

\subsection{Artificial data}
\begin{figure}
\centering  
\includegraphics[width=0.55\textwidth]{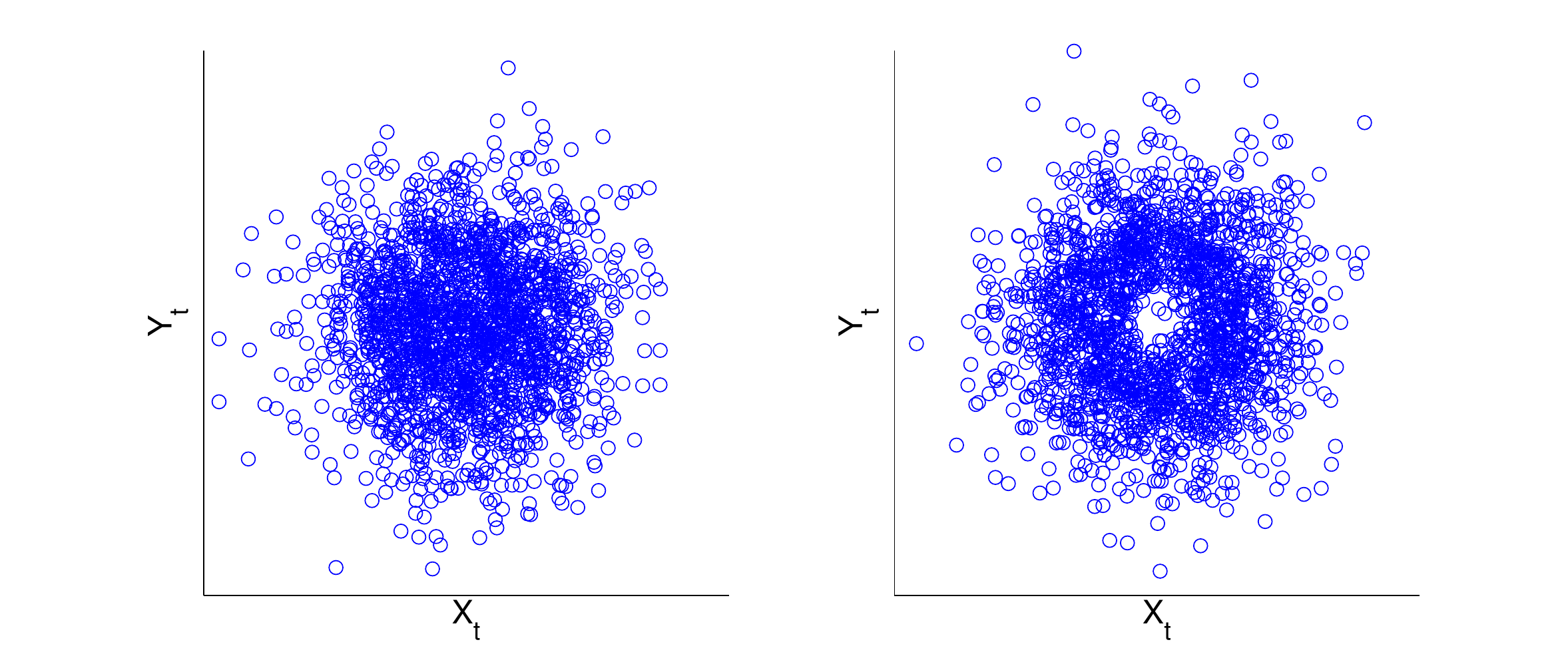}
\caption{$X_t$ and $Y_t$, described in the Experiment \ref{sec:tp}, with extinction rates $50 \%$ (left) and $99.8 \%$ (right), respectively.}
\label{fig:noise}
\end{figure}

\paragraph{Non-linear dependence.}
\label{sec:tp} 
We investigate two dependent, autoregressive random processes $X_t$,$Y_t$, specified by
\begin{equation}
\label{eq:AR1}
X_t = a X_{t-1} + \epsilon_t
\quad
Y_t = a Y_{t-1} + \eta_t,
\end{equation}
with an autoregressive component $a$. The coupling of the processes is a result of the dependence in the  innovations $\epsilon_t,\eta_t$. These $\epsilon_t,\eta_t$ are drawn from an Extinct  Gaussian distribution, defined in Algorithm~\ref{alg:example}. The parameter $p$ (called extinction rate) controls how often a point drawn form a ball $B(0,r)$ dies off. According to  Algorithm~\ref{alg:example}, the probability of seeing a point inside the ball $B(0,r)$ is different than for a two dimensional Gaussian $N(\mathbf 0 ,Id)$. On the other hand, as $p$ goes to zero, the Extinct Gaussian converges in distribution to $N(\mathbf 0,Id)$. Figure~\ref{fig:noise} illustrates the joint distribution of $X_t,Y_t$. The left scatter plot in Figure~\ref{fig:noise} presents $X_t$ and $Y_t$ generated with an extinction rate of  $50  \%$, while the right hand plot is generated with an extinction rate of $99.87 \%$.  Processes used in this experiment had an autoregressive component of $0.2$, and the radius of the innovation process was $1$.

\begin{figure}[t]
\centering  
\includegraphics[width=0.5\textwidth]{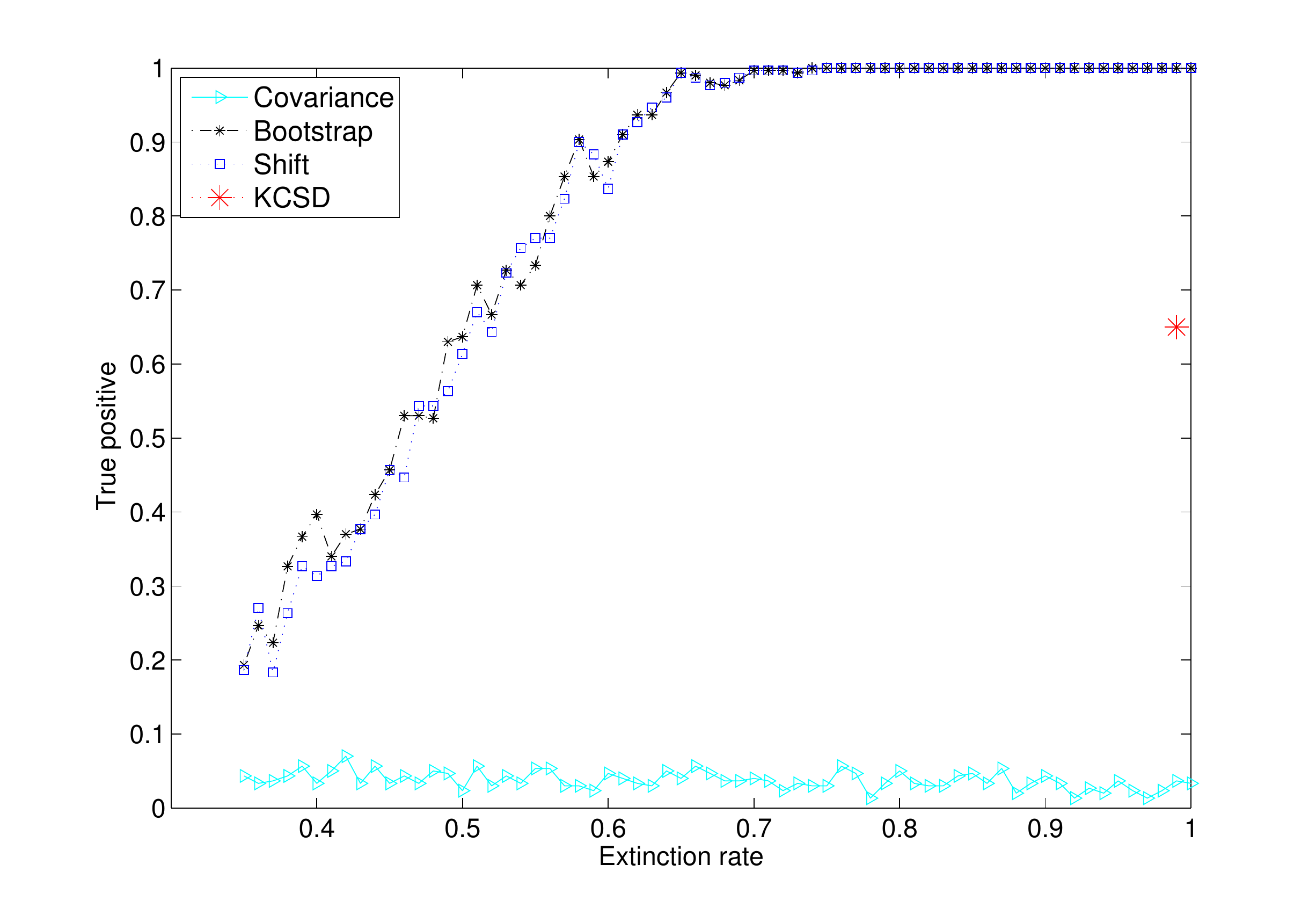}
\caption{True positive rate for the Shift HSIC, the Bootstrap HSIC and correlation based test: sample size $1200$, results averaged over $300$ repetitions.  The red star shows  KCSD performance at $4\times$ the HSIC sample size; see Section \ref{sec:tp} for details.}
\label{fig:tp2}
\end{figure}

Figure~\ref{fig:tp2} compares the power of the Shift HSIC test and the correlation test. The $X$ axis represents an extinction rate, while the $Y$ axis shows the true positive rate.  Shift HSIC  is capable of detecting non-linear dependence between $X_t$ and $Y_t$, which is missed by linear correlation. The red star depicts performance of the KCSD algorithm developed by \citet{besserve_statistical_2013}, with parameters tuned by its authors: note that this result required using four times as many data points as HSIC.

\begin{algorithm}[tb]
   \caption{Generate innovations}
   \label{alg:example}
\begin{algorithmic}
   \STATE {\bfseries Input:} extinction rate $0 \leq p \leq 1$, radius $r$.
   \REPEAT
   \STATE Initialize $\eta_t,\epsilon_t$ to $N(0,1)$ and $d$ to a number  uniformly distributed on $[0,1]$ .
   \IF{$\eta_t^2 +\epsilon_t^2  > r^2 $ \OR $d>p$} 
   \STATE return $\eta_t,\epsilon_t$
   \ENDIF
   \UNTIL{true}
\end{algorithmic}
\end{algorithm}

\begin{figure}[t]
\centering  
\includegraphics[width=0.5\textwidth]{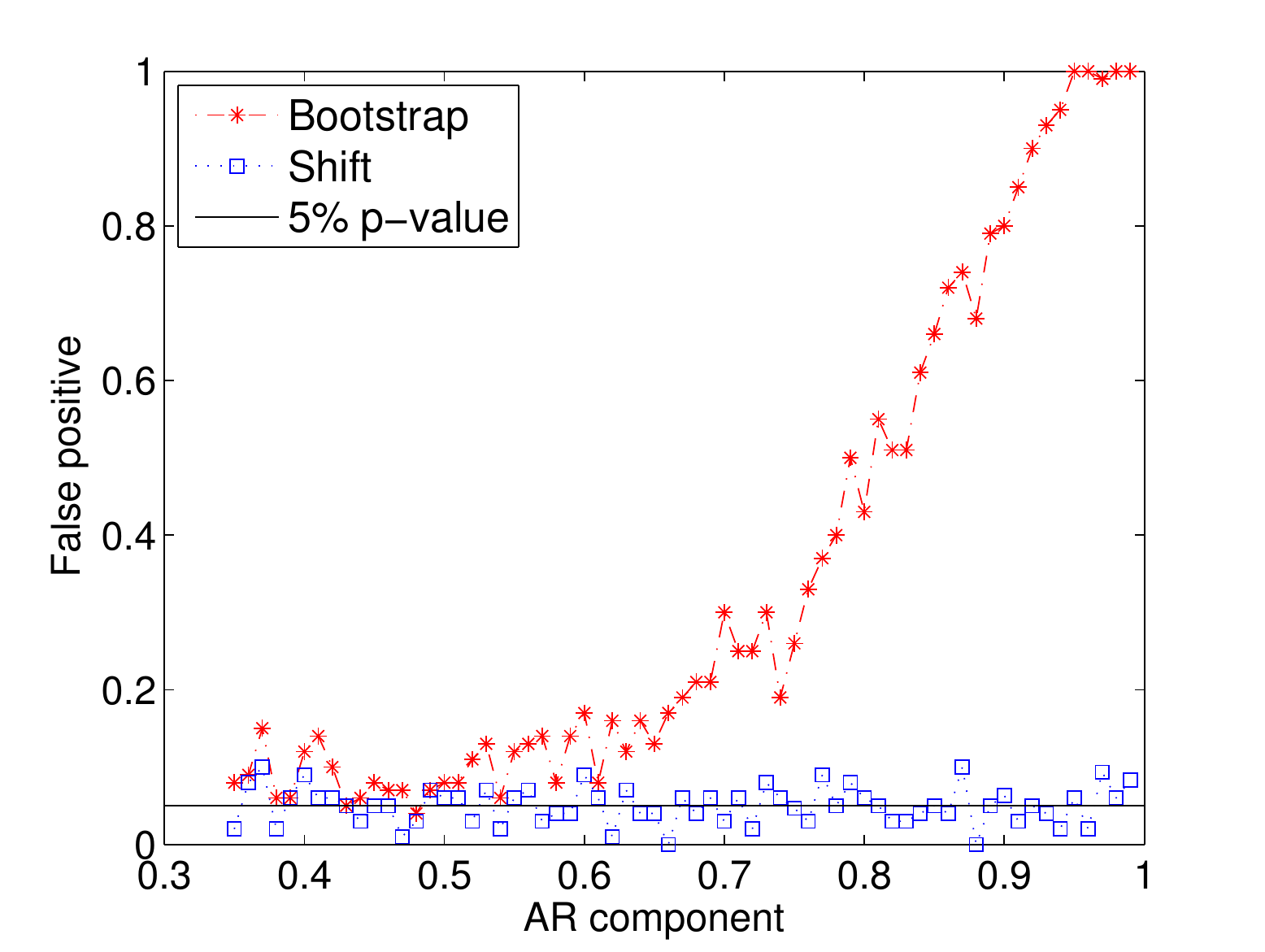}
\caption{False positive rate for the Shift HSIC and the Bootstrap HSIC. The sample size was $1200$, and results were averaged over $300$ repetitions.}
\label{fig:AB}
\end{figure}

\vspace{-2mm}
\paragraph{False positive rates.} 
\label{ex:ar1}
We next investigate the rate of false positives for Shift HSIC and Bootstrap HSIC on independent copies of the $AR(1)$ processes used in the previous experiment. To generate independent processes,  we first sampled two pairs $(X_t,Y_t)$, $(X_t',Y_t')$ of time series using \eqref{eq:AR1}, and then  constructed $Z$ by  taking $X$ from the first pair and $Y$ from the second, i.e., $Z_t=(X_t,Y_t')$. We set an extinction rate to $50\%$. \footnote{As a reviewer pointed out, the example for the FP rates can be simplified, however we decided to be consistent with the marginal distribution of $X_t$,$Y_t$ across the experiments.} The AR component $a$ in the model \eqref{eq:AR1} controls the memory of a processes - the larger this component, the longer the memory. We  performed the  Shift HSIC and the Bootstrap HSIC tests on $Z_t$ generated under $\mathbf{H_0}$  with different AR components. Figure~\ref{fig:AB} illustrates the results of this experiment. The $X$ axis is indexed by the AR component and $Y$ axis shows the FP rate. As the temporal dependence increases, the Bootstrap HSIC incorrectly gives an increasing number of false positives: thus, it cannot be relied on to detect dependence in time series. The Shift HSIC false positive rate remains at the targeted $5\%$ p-value level.


\subsection{Forex data}\label{sec:forex}
We use Foreign Exchange Market quotes to evaluate  Shift HSIC performance on the real life data. Practitioners point out that  forex time series are noisy and hard to handle, especially at low granulations (smaller then 15 minutes). We decided to work with forex time series to show that  Shift HSIC can detect dependence even on such a difficult dataset.    
The forex time series were granulated to obtain two minute sampling (the granulation function returned the last price in the two minute window). Using the test of \citet{diks_nonparametric_2005}, we checked that serial dependence of the differentiated  time series decays fast enough to satisfy the  assumed mixing conditions (by a differentiated time series, we refer to $(X_t -X_{t-1})_{t \in \mathbf{N}}$). The choice of the pairs and  trading day (21st January 2013) were arbitrary.

\paragraph{Instantaneous coupling and causal effect.}
Having one Australian dollar we may obtain a quantity of Yen in two ways, either by using AUD/JPY exchange rate explicitly or by buying  Canadian dollars  and then selling them at the  CAD/JPY rate. Let $X_t$ be a differentiated  AUD/JPY exchange rate and $Y_t$ be a differentiated product of exchange rates AUD/CAD$\times$CAD/JPY. We will investigate the relation between these two. Common sense dictates that $Y_t$ should behave similarly to $X_t$. After examining the cross-correlation of $X_t$ and $Y_t$, we propose a simple regression model to describe the interaction between the signals,
\[
\hat Y_t =  a_0X_t + a_1 X_{t-1} + \cdots + a_6 X_{t-6}.
\]
We fit the model and see that $a_0=0.97$, and the remaining coefficients are not bigger then $0.06$ in  absolute value. This suggest that most of the dependence is explained by an instantaneous coupling. We further investigate the cross-correlation between residuals $R_t = Y_t - \hat Y_t$ and $X_t$. We observe  no significant correlations in the first 30 lags.

\begin{figure}
\centering  
  \includegraphics[width=0.5\textwidth]{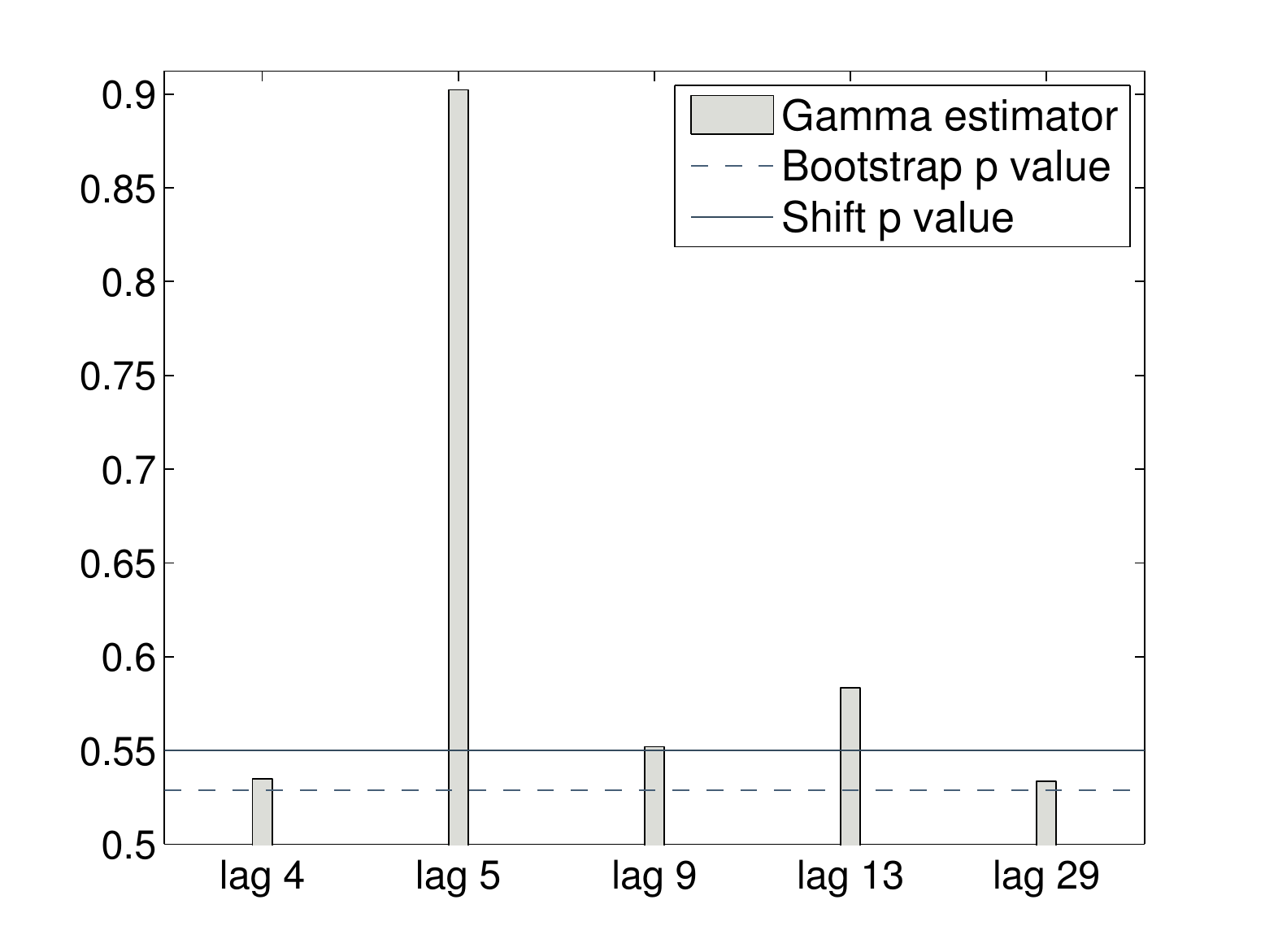}
\caption{Instantaneous coupling. Results for 720 samples, null threshold of Shift HSIC used $300$ lags in range $100-400$.}
\label{fig:pValues}
\end{figure}

Next we investigate dependence of residuals with lagged values of the explanatory variables, i.e., $R_t$ with $X_{t-k}$ for $k \in (0,\cdots,30)$.  After calculating p-values using the Bootstrap HSIC and the Shift HSIC, we discover dependence only at lags $4$, $5$, $9$, $13$ and $29$, as presented in the Figure \ref{fig:pValues}. Lack of the dependence at lag zero suggests that the linear model for coupling is reasonable. However, both the Bootstrap HSIC and the Shift HSIC support the hypothesis that there is a strong relation at lag $5$, which is not explained well by the linear model. 

The questions remains whether test statistics at lags $4$, $9$, $13$ and $29$ indicate further model misspecification. Under $\mathbf{H_0}$, at a significance level $94\%$, we expect 1.8 out of 30 statistics to be higher than the 94th percentile. Excluding the statistic at lag $5$, the  Shift HSIC test reports two statistics above this percentile, while Bootstrap HSIC reports four. Should the statistics at the  different lags be independent from each other, the probabilities of seeing two and four statistics above the percentile are respectively $25 \%$ and $6\%$. Shift HSIC  indicates that the model fits the data well, while the Bootstrap HSIC suggests that some non-linear dependencies remain unexplained.

\begin{figure}
\centering  
\includegraphics[width=0.5\textwidth]{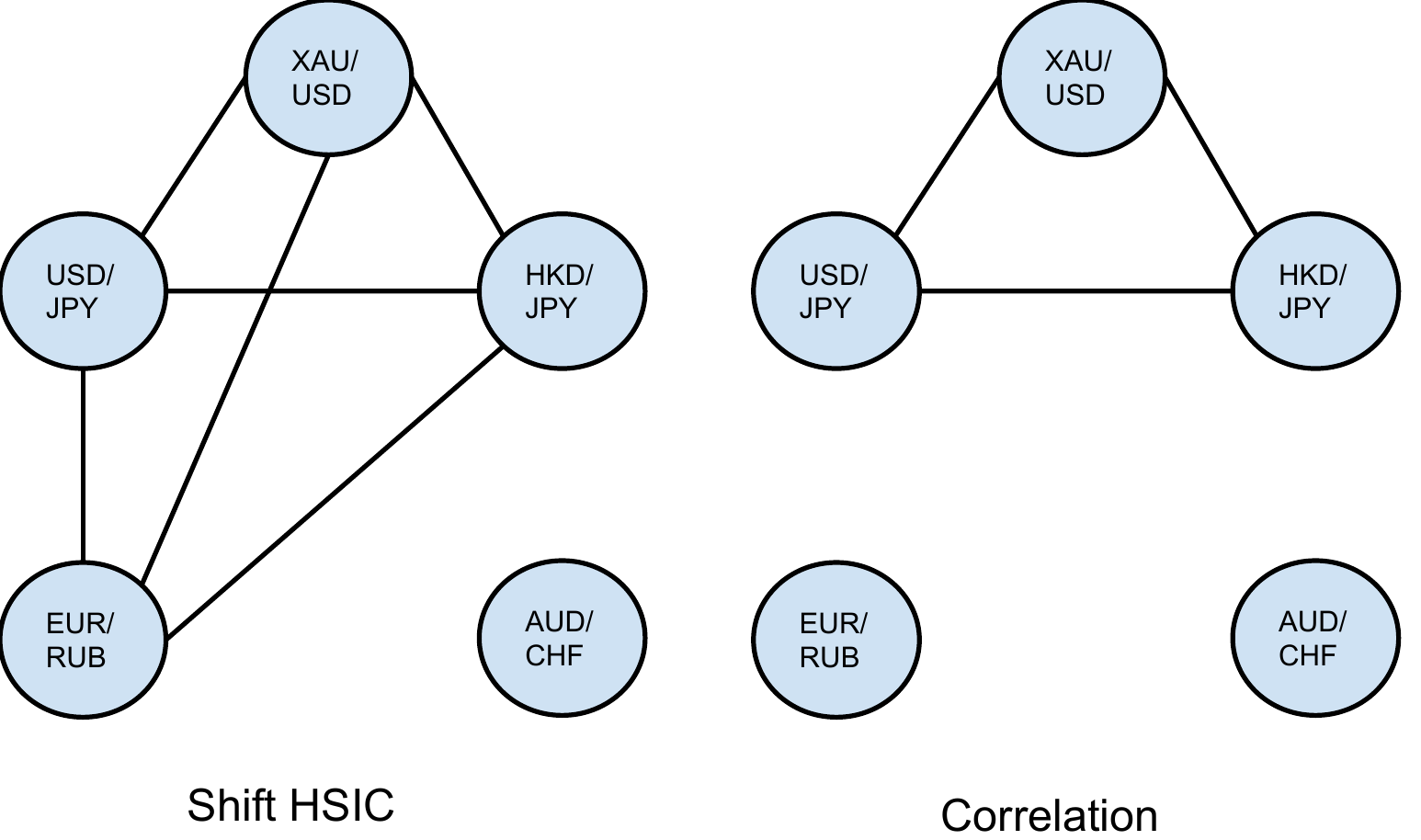}
\caption{Differences between the dependence structure on the forex revealed by the Shift HSIC and covariance. Parameter settings are as in Figure \ref{fig:pValues}.}
\label{fig:Shift}
\end{figure}

\vspace{-2mm}
\paragraph{Dependence structure.} 
The data are  five currency pairs. A correlation based independence  test, and the Shift HSIC test,  were performed on each pair of  currencies. The dependencies revealed by these tests are depicted  in  Figure~\ref{fig:Shift} - nodes represent  the time series and edges represent dependence. Shift HSIC  reveals a strong coupling between EUR/RUB and USD/JPY, HKD/JPY and XAU/USD that was not found by simple correlation. All edges revealed by Shift HSIC have p-values at most at level $0.03$ - clearly, the Shift HSIC managed to find a strong non-linear dependence. Note that the obtained graphs are cliques.

\vspace{-2mm}
\section{Proofs}\label{sec:proofs}
 A $U$-statistic of a $k$-argument, symmetric function $f$, is written
 \begin{equation*}
 \label{def:Ustat}
  U(f,Z) = {n \choose k}^{-1} \sum_{1 \leq i_1 < \cdots < i_k \leq n} f(Z_{i_1},...,Z_{i_k}).
 \end{equation*}
A  decomposition due to Hoeffding allows us to decompose this U-statistic into a sum of $U$-statistics of canonical functions, $ U(h,Z) =  \sum_{k=1}^{l} {l \choose k} U(h_k,Z)$,
where  $h_k(z_1,...,z_l)$ are  components of the  decomposition. According to \citet[section 5.1.5]{serfling_approximation_2002}, each of  $h_1$,$h_2$,$h_3$,$h_4$ is symmetric and canonical.  Note that $h_k$ is defined using independent samples $Z^*$ - this is because the CLT or LLN state that U-statistics or V-statistics of mixing processes converge to their expected value taken with respect to independent copies, i.e., $Z^*$. Under $\mathbf{H}_0$, $h_1$ is equal to zero everywhere and $h_2 =  \frac 1 6 s$, where these results were obtained by \citet{gretton_kernel_2008}.\footnote{The second result is hard to locate - it is in appendix A.2, text between equations 12 and 13} See supplementary material for  details.  

In order to characterize $U(h,Z)$,  we show that under null hypothesis $U(h_2,Z)$ converges to a random variable, and both $U(h_3,Z)$,$U(h_4,Z)$ converge  to zero in a probability (the latter proof can be found in the supplementary material). Bellow we characterise $U(h_2,Z)$ convergence. 
\newcommand{\tp}{\mathbf{t}}
\begin{lemma}
\label{th:h2convergance}
Under assumptions of Lemma \ref{lem:getReady}, 
\[
\lim_{n \to \infty } n \cdot U(h_2,Z) \stackrel{D}{=} \frac 1 6  \sum_{i_1}^{\infty} \lambda_{i_1} (\tau_{i_1}^2-1).
\]
\end{lemma} 

\begin{proof}
First recall that under null hypothesis $h_2 =  \frac 1 6 s$. 
We will check the conditions of \citep[Theorem 1]{i._s._borisov_orthogonal_2009} (also available in the supplementary).\\
First, from Mercer's Theorem \citep[Corollary 3.5]{steinwart_mercers_2012},  we deduce that the $h_2$ coefficients in $L_2(\mathbf{Z}, \mathcal{B}_\mathbf{Z},P_\mathbf{Z})$ are absolutely summable. In the supplementary material we show that $\ev e_i(Z_1^*)=0$. \\
Recall the assumptions of Lemma \ref{lem:getReady}.  If $\textbf{A}$ holds then $\sum_{k=1}^{\infty} \phi(k)^{\frac{1}{2}} < \infty$ and $\sup_i \ev | e_i(X_1)|^2 =1 < \infty$ ($e_i$ is an orthonormal eigenfunction).
Finally, if $\textbf{B}$ holds then the process $Z_t$ is $\alpha$-mixing. The remaining assumptions concerning uniform mixing in \citet{i._s._borisov_orthogonal_2009} are exactly the same as in this lemma.\\
\end{proof}
\vspace{-5mm}
\subsection{Main body proofs}\label{sec:mainProofs}
 
\vspace{-2mm}
\begin{proof}
(Lemma \ref{lem:getReady}) We use the fact that $h_2$ is equal to $s$ up to scaling ($6 U(h_2,Z) =  U(s,Z)$), and Lemma \ref{th:h2convergance}, to see that $n U(s,Z) \stackrel{D}{\to} \sum_{i}^{\infty} \lambda_{i} (\tau_{i}^2-1)$. Since $\ev s(Z_t,Z_t) = \ev \sum_{i=1}^{\infty} \lambda_i e_i(Z_t)^2  = \sum_{i=1}^{\infty} \lambda_i$, then by the LLN for mixing processes,
\begin{equation}
\label{eq:sumOfEigen}
\lim_{n \to \infty} \frac 1 n \sum_{i}^n s(Z_i,Z_i) \stackrel{P}{=} \sum_{i=1}^{\infty} \lambda_i. 
\end{equation} 
We use a relationship between $U$ and $V$ statistics,
\begin{equation*}
\begin{split}
\lim_{n \to \infty}& n V(s,Z)  \stackrel{D}{=}  \lim_{n \to \infty} n U(s,Z) + \lim_{n \to \infty} \frac 1 n \sum_{i}^n s(Z_i,Z_i) \\  
 &\stackrel{D}{=} \sum_{i=1}^{\infty} \lambda_i + \sum_{i}^{\infty} \lambda_{i} (\tau_{i}^2-1) \stackrel{D}{=} \sum_{i}^{\infty} \lambda_{i} \tau_{i}^2.
\end{split}
\end{equation*}
\end{proof}
\vspace{-2mm}
\begin{proof}
 (Theorem \ref{th:main}) We operate under the null hypothesis. Recall that  $U(h,Z)$ can be decomposed as $U(h,Z) = \sum_{k=1}^{4} {4 \choose k} U(h_k,Z)$. Here $h_1 \equiv 0$. We show in the supplementary material that $U(h_3,Z)$ and $U(h_4,Z)$ tend to zero in probability. 
From Lemma \ref{th:h2convergance}, 
\begin{equation}
\label{eq:one}
\lim_{n \to \infty} n  U(h,Z) \stackrel{D}{=} \lim_{n \to \infty} n U(s,Z) \stackrel{D}{=} \sum_{i}^{\infty} \lambda_{i} (\tau_{i}^2-1).
\end{equation} 
We define an auxiliary symmetric function $w$, 
 \begin{equation*}
 \begin{split}
   w(z_1,z_2,z_3) &= h(z_1,z_1,z_2,z_3)+ h(z_1,z_2,z_2,z_3) \\ 
   &+ h(z_1,z_2,z_3,z_3) +h(z_1,z_1,z_3,z_2)  \\
   &+ h(z_3,z_2,z_2,z_1)+ h(z_2,z_1,z_3,z_3).
 \end{split}
 \end{equation*}
  It is obvious that $\ev w(Z_1^*,Z_2^*,Z_3^*)$ $=$ $6\ev h(Z_1^*,Z_1^*,Z_2^*,Z_3^*)$. We consider the difference between the unnormalized $V$ and $U$ statistics,
   \begin{equation*}
  \begin{split}
  S_n  =&  \sum_{1 \leq i_1,i_2,i_3,i_4 \leq n} h(Z_{i_1},...,Z_{i_4})-\sum_{i \in C_4}  h(Z_{i_1},...,Z_{i_4}),\\        
  \end{split}
 \end{equation*} 
where $\sum_{i \in C_m}$ denotes summation over all $n \choose m$ combinations of $m$ distinct elements $\{ i_1, \cdots ,i_m\}$ from $\{1, \cdots ,n\}$. The difference is equal to the sum over $4$-tuples with at least one pair of equal elements.  We can choose such tuples in $\binom 4 2 = 6$ ways. Observe that $w$ covers the choice of all these six tuples. Since for any $z_1,z_2 \in \mathbf{Z}$, $h(z_1,z_1,z_1,z_2)=0$, then $w$ is zero whenever more than two indices are equal. Therefore we can sum $w$ over distinct  indices $z_1,z_2,z_3$,
 \begin{equation*} 
  S_n  = \sum_{i \in C_3}  w(Z_{i_1},Z_{i_2},Z_{i_3}).
 \end{equation*}
 We see that $S_n$ is almost a $U$-statistic ($U(w,Z)$). By the CLT for $U$-statistics from \citet{denker_u-statistics_1983}, Theorem 1(c), we obtain
 \begin{equation*}
  \lim_{n \to \infty} \frac 1 {n(n-1)(n-2)}S_n \stackrel{P}{=}  6 \ev h(Z_1^*,Z_1^*,Z_2^*,Z_3^*).
 \end{equation*}
  On the other hand, via the relation $h_2 =  \frac 1 6 s$ and the  $h_2$ definition, we get $\ev s(Z_1^*,Z_1^*)  = 6 \ev h(Z_1^*,Z_1^*,Z_2^*,Z_3^*)$, and therefore  
 \begin{equation}
   \label{eq:fancyLimit}
   \lim_{n \to \infty} \frac 1 {n(n-1)(n-2)}S_n  \stackrel{P}{=} \lim_{n \to \infty} \frac 1 n \sum_{i}^n s(Z_i,Z_i) .
 \end{equation}
 Finally, we rewrite $S_n$ as
 \begin{equation*}
  \begin{split}
  \sum_{  1 \leq i_1,i_2,i_3,i_4 \leq n}&  h(Z_{i_1},...,Z_{i_4})= S_n +  \sum_{ i \in C_4} h(Z_{i_1},...,Z_{i_4}). \\
  \end{split}
 \end{equation*}  
We normalize  by $\frac{1}{n(n-1)(n-2)}$, and take the limit in $n$,
 \begin{equation*}
  \begin{split}     
   \lim_{n \to \infty}& \frac {n^4} {n(n-1)(n-2)} V(h,Z) \stackrel{D}{=}\\ 
     &=  \lim_{n \to \infty} \left( \frac 1 {n(n-1)(n-2)} S_n +   (n-4)U(h,Z) \right). \\ 
 \end{split}
 \end{equation*}    
 We substitute (\ref{eq:fancyLimit}) and (\ref{eq:one}) on the right hand side, and use equation (\ref{eq:sumOfEigen}) from Lemma \ref{lem:getReady} to replace $\lim_{n \to \infty}  \frac 1 n \sum_{i}^n s(Z_i,Z_i)$ with $\sum_{i=1}^{\infty} \lambda_i$, yielding
 \begin{equation*}
 \begin{split}     
 \lim_{n \to \infty}& n \cdot V(h,Z) \stackrel{D}{=} \\  
     &\stackrel{D}{=}\lim_{n \to \infty}  \frac 1 n \sum_{i}^n s(Z_i,Z_i) +  \lim_{n \to \infty}   \frac 1 n \sum_{i, j}^n s(Z_i,Z_j) \stackrel{D}{=}\\
     &\stackrel{D}{=} \sum_{i=1}^{\infty} \lambda_i + \sum_{i}^{\infty} \lambda_{i} (\tau_{i}^2-1) \stackrel{D}{=} \sum_{i}^{\infty} \lambda_{i} \tau_{i}^2.
\end{split}
 \end{equation*}         
\end{proof}
\vspace{-2mm}
\begin{proof}
 (Theorem \ref{th:main2})  If the null hypothesis does not hold, then $\gamma >0$  \cite{gretton_measuring_2005}. In this case $h$ is nondegenerate, and we can use \citet[Theorem 1(c)]{denker_u-statistics_1983} to see that $\frac{\sqrt{n}}{4\ \sqrt{\sigma}}( V(h,Z)-\gamma) \sim N(0,1)$, where $\sigma$ is  finite (see the note below Theorem 1 of \citep{denker_u-statistics_1983}, stating that in case (c) $\sigma^2$ is  finite, and the note above Theorem 1 stating that $\sigma^2= \lim_{n \to \infty} n^{-1} \sigma_n^2$  ).
\end{proof}

\vspace{-2mm}
\begin{proof}
 (Lemma \ref{lem:basisRelation}) We use Lemma 1 and Theorem 4 from \citet{gretton_measuring_2005} to show that $\ev h(Z_1^*,Z_2^*,Z_3^*,Z_4^*)=0$ iff $(X_1^*,Y_1^*)$ has a product distribution. Since $Z_1^* \stackrel{D}{=}Z_1$ and $Z_t \stackrel{D}{=}Z_1$, we infer that $X_t$ is independent from $Y_t$ iff $\ev h(Z_1^*,Z_2^*,Z_3^*,Z_4^*)=0$.   
\end{proof}
\small
\vspace{-2mm}
\paragraph{Acknowledgements}
The authors thank the reviewers and colleagues for helpful feedback, especially  M. Skomra, D. Toczydlowska, and A. Zaremba.

\bibliography{dep_short}
\normalsize
\bibliographystyle{icml2014}

\appendix
\onecolumn

\section{A Kernel Independence Test for Random Processes - Supplementary}
The sections in the supplementary material are in the same order those in the article. In particular, the $n$-th reference to the supplementary in the  article is $n$-th subsection in the supplementary material.  

The arXiv version of the report and supplementary may be found at: \url{http://arxiv.org/abs/1402.4501}

Before we start, we cite \citep[Lemma 1]{yoshihara_limiting_1976}, which will be used below.

\begin{lemma}
\label{lem:divLemma} \cite{yoshihara_limiting_1976}
Let $(Z_t)_{t \in \mathbb{N}+}$ be an absolutely regular process with a mixing coefficient $(\beta(n))_{n \in \mathbb{N}+}$. Let $(t_1,t_2, \cdot, t_l)$ be a non-decreasing $l$-tuple, and let $j$ be an integer such that $2 \leq j \leq l$. Finally, let $g:\mathbb{R}^l \to \mathbb{R}$ be a measurable function satisfying
\begin{equation*}
\label{eq:limitedMoments}
\left( \ev |g(Z_{t_1}, \cdots , Z_{t_l})|^{1+\delta} \right) \leq M 
\end{equation*}
for some $\delta>0,M>0$. Then
\begin{equation*}
\left|\ev  g(Z_{t_1}, \cdots , Z_{t_l}) - \ev g(Z_{t_1}, \cdots , Z_{t_{j-1}}, Z_{t_j}^*, \cdots , Z_{t_l}^* ) \right| 
\leq 4M^{\frac 1 {1+\delta}} \beta(t_j-t_{j-1})^{\frac \delta {1+\delta}}.
\end{equation*}
\end{lemma}

Note that if a function $g$ is symmetric, then we can always reorder its arguments if necessary.

\subsection{Testing procedure - convergence of $S_n$ from equation $(5)$.}
Let $\pi$ be a permutation drawn from a uniform distribution over the set of $n$-element permutations. We will prove that the random variable 
\[
Q_n = \sum_{i=1}^n \frac 1 {i^{\frac 3 2}} \frac 1 {|\pi(1) -\pi(i)|^{\frac 3 2}} 
\]
converges to zero in probability at rate $O(n^{-1})$. Since $0\leq S_n \leq Q_n$, then $S_n$ converges to zero in probability at the same rate.

\begin{lemma}
\label{lem:evSum}
$\ev |\pi(1) - \pi(i)|^{-{\frac 3 2}} = O(n^{-1})$.  
\end{lemma}
\begin{proof}
Let $j$ be a positive integer smaller than $n$. Observe that the sum $\sum_{i}^{n} |j- i|^{-{\frac 3 2}}$ is finite, 
\begin{equation}
\label{eq:niceSum}
 \sum_{i}^{n} |j- i|^{-{\frac 3 2}} \leq 2 \sum_{i}^{n} i^{-{\frac 3 2}} \leq 2 \zeta\left({\frac 3 2}\right),
\end{equation}

where $\zeta(\cdot)$ is the Riemann  zeta function. Now expand the expected value $\ev |\pi(1) - \pi(i)|^{-{\frac 3 2}}$ using a conditional expected value,
\begin{equation}
\begin{split}
\ev& |\pi(1) - \pi(i)|^{-{\frac 3 2}}  = \ev (\ev |j - \pi(i)|^{-{\frac 3 2}} | \pi(1)=j)    = \sum_{j=1}^n \frac 1 n (\ev |j - \pi(i)|^{-{\frac 3 2}} | \pi(1)=j) =\\ 
&=  \sum_{j=1}^n \frac 1 n \sum_{j \neq 1}^n \frac 1 {n-1} |j - i|^{-{\frac 3 2}}  \leq  \frac 1 {n(n-1)} \sum_{j=1}^n 2 \zeta\left({\frac 3 2}\right) = 2 \zeta\left({\frac 3 2}\right) \frac  1 {n-1}.    
\end{split}
\end{equation}
\end{proof}
 
\begin{lemma}
\label{lem:sumCovariance}
If $k \neq j$ are  positive integers smaller than $n$, then  
$$
\ev |\pi(k) -\pi(1)|^{-\frac 3 2} |\pi(j) -\pi(1)|^{-\frac 3 2} = O\left( \frac 1 {n^2}\right)$$ 
\end{lemma}
\begin{proof}
We   use the inequality \eqref{eq:niceSum} and properties of a conditional expected value.
\begin{equation}
\begin{split}
\ev& |\pi(k) -\pi(1)|^{-\frac 3 2} |\pi(j) -\pi(1)|^{-{\frac 3 2}} =  \ev \left( \ev |\pi(k) -a|^{-{\frac 3 2}} |\pi(j) -a|^{-{\frac 3 2}} \big|\pi(1)=a \right)   \\
&= \frac 1 n \sum_{a=1}^n \left( \ev |\pi(k) -a|^{-\frac 3 2} |\pi(j) -a|^{-\frac 3 2} \big| \pi(1)=a \right) \\
 &= \frac 6 {n(n-1)(n-2)} \sum_{a \neq b, a \neq c,b \neq c}^n \frac 1 {|b-a|^{\frac 3 2}} \frac 1 {|c-a|^{\frac 3 2}} \\
 & \leq \frac 1 {n(n-1)(n-2)} \sum_{a \neq b}^n \frac {2 \zeta\left({\frac 3 2}\right)} {|b-a|^{\frac 3 2}}  \\
&\leq
   \frac 1 {n(n-1)(n-2)} \sum_{a}^n 4 \zeta\left({\frac 3 2}\right)^2 \\
&=
 \frac 1 {(n-1)(n-2)}4 \zeta\left({\frac 3 2}\right)^2 \\
&= O\left( \frac 1 {n^2}\right)
\end{split}
\end{equation}

\end{proof}

\begin{lemma}
$Q_n$ converges to zero in probability. The convergence rate is $\frac 1 n$.
\end{lemma}
\begin{proof}
First, using Lemma \ref{lem:evSum} , we compute the expected value of $Q_n$
\[
\ev Q_n = \ev \sum_{i=1}^n \frac 1 {i^{\frac 3 2}} \frac 1 {|\pi(1) -\pi(i)|^{\frac 3 2}} = \sum_{i=1}^n \frac 1 {i^{\frac 3 2}} \ev  \frac 1 {|\pi(1) -\pi(i)|^{\frac 3 2}} \leq \sum_{i=1}^n \frac 1 {i^{\frac 3 2}} \frac 1 n C \leq \frac 1 n C \zeta\left({\frac 3 2}\right) = O\left(\frac 1 n\right).
\]
Next, using Lemma \ref{lem:sumCovariance}, we compute the second moment  
\begin{equation}
\begin{split}
\ev & \left( \sum_{k=1}^n \frac 1 {k^{\frac 3 2}} \frac 1 {|\pi(k) -\pi(1)|^{\frac 3 2}} \right) \left( \sum_{j=1}^n \frac 1 {j^{\frac 3 2}} \frac 1 {|\pi(j) -\pi(1)|^{\frac 3 2}} \right) \\
& \leq \ev \left( C \frac 1 {n^2} \sum_{k \neq j}^n \frac 1 {k^{\frac 3 2}} \frac 1 {j^{\frac 3 2}} +  \sum_{k=1}^n \frac 1 {k^3} \frac 1 {|\pi(k) -\pi(1)|^3} \right) \\
& \leq  C \frac 1 {n^2} \zeta\left({\frac 3 2}\right)^2 + C' \frac 1 n \zeta(3) = O\left(\frac 1 n\right).
\end{split}
\end{equation}

Using the Chebyshev's inequality we obtain the required  result.
\end{proof}

\subsection{Testing procedure - Shift HSIC samples from the right distribution}
We will investigate the value of the $V$-statistic for a shifted process i.e. $n V(h,Z^k)$. 


\paragraph{Null hypothesis holds.} If the null hypothesis holds, then  $X_t$ and $Y_{t+k}$ are independent for any $k$. To see this, suppose that there exists $k$ for which $X_t$ and $Y_{t+k}$ are dependent (the processes are stationary, so this is true for all $t$). The observation $X_t$ depends on its past values: in particular, $X_{t-k}$ is a parent of $X_t$. If  in addition $X_{t-k} \rightarrow Y_t$, then $Y_t$ and $X_t$ will be dependent, as they share a parent.

We will use this fact to  show that the $n V(h,Z^k)$ has the same distribution as the $n V(h,Z)$. Recall the covariance structure of $n V(h,Z)$ from Theorem $\ref{th:main}$,
\begin{equation}
\label{eq:covStrucuture}
\begin{split}
 \ev \tau_a \tau_b &= \ev e_a(Z_1) e_b(Z_1) + \sum_{j=1}^{\infty} \left[ \ev e_a(Z_1)e_b(Z_{j+1}) + \ev e_b(Z_1)e_a(Z_{j+1}) \right].
 \end{split}
\end{equation}
We represent $e_a$ and $e_b$ as $e_a(z) = e^X_u(x)e^Y_o(y)$, $e_b(z) = e^X_i(x)e^Y_p(y)$. This represents  a decomposition  of the basis of $\mathbf Z$ into basis of $\mathbf X,\mathbf Y$, respectively. Consider one of the above infinite sums with $Y_t$ replaced with the shifted process $S^k_t$,
\begin{equation}
 \begin{split}
  T_n = \sum_{j=1}^{n}  \ev e_a(Z^k_1)e_b(Z^k_{j+1})  = \sum_{j=1}^{n}  \ev e^X_u(X_1)e^X_i(X_{j+1}) e^Y_o(S^k_1)e^Y_p(S^k_{j+1}).
 \end{split}
\end{equation}
We obtain the following covariance structure  for  $n V(h,Z^k)$,
\[
\begin{split}
  T_n&  = \sum_{j=1}^{n}  \ev e^X_u(X_1)e^X_i(X_{j+1}) \ev  e^Y_o(S^k_1)e^Y_p(S^k_{j+1}) \\ 
 &=\sum_{j=1}^{n-k-1}  \ev e^X_u(X_1)e^X_i(X_{j+1}) e^Y_o(Y_{1+k})e^Y_p(Y_{j+1+k}) + \sum_{j=n-k}^{n}  \ev e^X_u(X_1)e^X_i(X_{j+1})  e^Y_o(Y_{1 + k})e^Y_p(Y_{1 + n-k-j})  \\
 &=\sum_{j=1}^{n-k-1}  \ev e^X_u(X_1)e^X_i(X_{j+1}) \ev e^Y_o(Y_{1+k})e^Y_p(Y_{j+1+k}) + \sum_{j=n-k}^{n}  \ev e^X_u(X_1)e^X_i(X_{j+1})  \ev e^Y_o(Y_{1 + k})e^Y_p(Y_{1 + n-k-j})  \\
 &\leq\sum_{j=1}^{n-k}  \ev e^X_u(X_1)e^X_i(X_{j+1}) e^Y_o(Y_1)e^Y_p(Y_{j+1}) +  O(k (n-k)^{-\frac 3 2}).  
\end{split}
\]
We have used the fact that $Y_t$ is stationary, $\ev e^Y_o(Y_{1+k})e^Y_p(Y_{j+1+k}) = e^Y_o(Y_1)e^Y_p(Y_{j+1})$ and  that the pairs $(X_1,X_{1+j})$, $(Y_{1+k},Y_{j+1+k})$ are independent (because $X_t$ and $Y_{t+k}$ are independent for all shifts $k$). For the second term,
\[
\sum_{j=n-k}^{n}  \ev e^X_u(X_1)e^X_i(X_{j+1})  \ev e^Y_o(Y_{1 + k})e^Y_p(Y_{1 + n-k-j}), 
\] 
we have used  covariance inequalities from \citet[section 1.2.2]{doukhan_markov_1994} and our bounds on mixing coefficients to obtain that when $j\ge n-k$, then $ \ev|e^X_u(X_1)e^X_i(X_{j+1})| \leq  (n-k)^{- \frac 3 2}$ (and by e.g. Holders inequality, $\ev e^Y_o(Y_{1 + k})e^Y_p(Y_{1 + (n-j)})$ is finite). The first component takes the form
\[
\sum_{j=1}^{n-k}  \ev e^X_u(X_1)e^X_i(X_{j+1}) e^Y_o(Y_1)e^Y_p(Y_{j+1}) = \sum_{j=1}^{n-k}  \ev e_a(Z_1)e_b(Z_{j+1}). 
\] 
Here
$\sum_{j=1}^{n-k} \ev e_a(Z_1)e_b(Z_{j+1})$ converges to $\sum_{j=1}^{\infty}  \ev e_a(Z_1)e_b(Z_{j+1})$ from equation \eqref{eq:covStrucuture}. Since $\ev e_a(Z^k_1) e_b(Z^k_1)=\ev e_a(Z_1) e_b(Z_1)$, the covariance structure from equation \eqref{eq:covStrucuture} is recovered.

\paragraph{Null hypothesis  does not hold.} In this case, the dependence between $X_t$ and $Y_{t+k}$ decreases as $k$ increases, since the mixing coefficients for each of the time series converges to zero. In the limit of large $k$ and $n$, the normalized $V$-statistic will converge to the null distribution, where $X_t$ and $Y_t$ are independent random processes. The proof of this result under the assumed mixing conditions, with suitable conditions on the increase of $k$ with $n$, is a topic of future work (the next two sections give an outline of the results that would need to be established for the shifted process).

  
\subsection{Proofs - Hoeffding decomposition}

The Hoeffding decomposition  \citep[e.g.][]{serfling_approximation_2002} allows us to decompose U-statistics into a sum of simpler $U$-statistics that can be easier to analyse. In the following section we will perform a Hoeffding decomposition of $U(h,Z)$ and investigate some of its properties. In the sequel we assume that  $k$ and $l$ are bounded kernels. For the $U$ statistic $U(h,Z)$, we call the function $h$ a {\em core}.

Any U-statistic can be written as a sum of V-statistics with degenerate cores. To show this, we define the auxiliary functions
\[
 g_c(z_1,...z_c) = \ev h(z_1,...,z_c,Z_{c+1}^*,...,Z_{m}^*)
\]
for each $c=1,...,m-1$ and put $g_m=h$.  

We assume the expected value of the core with respect to starred $\{Z_t\}$ is zero, i.e., $\ev h(Z_1^*,\cdots,Z_m^*) = 0$. The canonical functions that enable the core decomposition are 
\begin{equation*}
 \begin{split}
   h_1(z_1) &= g_1(z_1),\\
   h_2(z_1,z_2) &= g_2(z_1,z_2)  - h_1(z_1) - h_1(z_2), \\  
   h_3(z_1,z_2,z_3) &= g_3(z_1,z_2,z_3) - \sum_{1 \leq i < j \leq 3 } h_2(z_i,z_j) - \sum_{1\leq i \leq 3} h_1(z_i), \\ 
    & \vdots \\
   h_m(z_1,...,z_m) &= g_m(z_1,...,z_m) - \sum_{1 \leq i_1 < ...<i_{m-1} \leq m } h_{m-1}(z_{i_1},...,z_{i_{m-1}}) \\ 
    & - ... - \sum_{1 \leq i_1 < i_2 \leq m } h_2(z_{i_1},z_{i_2}) - \sum_{1 \leq i \leq m} h_1(z_i).\\
 \end{split}
 \end{equation*}
 We call these functions components of a core.

\begin{statement}
\label{stm:decomposition}
The  U-statistic of a core function $h$ can be written as a sum of  U-statistics with degenerate cores,
 \[
  U(h,Z) = U(h_m,Z) + \binom m 1 U(h_{m-1},Z) + ...+ \binom {m} {m-2} U(h_{2},Z) + \binom {m} {m-1} U(h_{1},Z).
 \]
\end{statement}

\begin{proof}
Recall that $\sum_{i \in C_m}$ denotes summation over all $n \choose m$ combinations of $m$ distinct elements $\{ i_1, \cdots ,i_m\}$ from $\{1, \cdots ,n\}$.
\begin{align*}
U(h,Z)& = \frac{1} {n^m} \sum_{i \in C_m} h(Z_{i_1},...,Z_{i_m})  \\
&= \frac{1} {n^m}  \sum_{i \in C_m}   \left( h_m(Z_1,...,Z_m) + \sum_{1 \leq j_1 < ...<j_{m-1} \leq m } h_{m-1}(Z_{j_{i_1}},...,Z_{i_{j_{m-1}}}) \right. \\ 
    &\quad \left. + ... + \sum_{1 \leq j_1 < j_2 \leq m } h_2(Z_{i_{j_1}},Z_{i_{j_2}}) + \sum_{1 \leq j \leq m} h_1(Z_{i_j}) \right) \\
    &= \frac{1} {n^m}  \sum_{i \in C_m}   h_m(Z_1,...,Z_m) +  \binom m 1 \frac{1} {n^{m-1}}  \sum_{i \in N^{m-1}} h_{m-1}(Z_{i_1},...,Z_{i_{m-1}}) + \\
    &\quad + ... + \binom {m} {m-2} \frac{1} {n^2}  \sum_{i \in N^2} h_2(Z_{i_1},Z_{i_2}) +  \binom {m} {m-1} \frac{1} {n} \sum_{i \in N} h_1(Z_i) \\
    &=U(h_m,Z) + \binom m 1 U(h_{m-1},Z) + ...+ \binom {m} {m-2} U(h_{2},Z) + \binom {m} {m-1} U(h_{1},Z).
\end{align*}
\end{proof}

\begin{lemma}
\label{lem:degenerate}
Under $\mathbf{H_0}$, $\forall z \in \mathbf{Z}$ $h_1(z) = 0$. 
\end{lemma}
\begin{proof}
 We use the shorthand notation $k(a,b) \equiv k(x_a,x_b)$, $l(a,b) \equiv l(y_a,y_b)$, such that
\begin{equation*}
h(z_a,z_b,z_c,z_d) = \frac{1}{4!} \sum_{\pi \in S_4} k(\pi_1,\pi_2)\left[ l(\pi_1,\pi_2) + l(\pi_3,\pi_4) - 2l(\pi_2,\pi_3) \right].
\end{equation*} 
Let us expand this expression. By using the  symmetry of $k$ and $l$, and writing the arguments in lexicographical order, we obtain
\begin{equation*}
\begin{split}
&h(z_a,z_b,z_c,z_d) =\\ 
&k(a,b) \left( l(a,b)+l(c,d)-2l(b,c)\right)+ k(a,b)\left(l(a,b)+l(c,d)-2l(b,d)\right) \\
&k(a,c) \left(l(a,c)+l(b,d)-2l(b,c)\right)+k(a,c)\left(l(a,c)+l(b,d)-2l(c,d)\right)+ \\
&k(a,d) \left(l(a,d)+l(b,c)-2l(b,d)\right)+k(a,d)\left(l(a,d)+l(b,c)-2l(c,d)\right)+\\
&k(a,b) \left(l(a,b)+l(c,d)-2l(a,c)\right)+k(a,b)\left(l(a,b)+l(c,d)-2l(a,d)\right)+\\
&k(b,c) \left(l(b,c)+l(a,d)-2l(a,c)\right)+k(b,c)\left(l(b,c)+l(a,d)-2l(c,d)\right)+\\
&k(b,d) \left(l(b,d)+l(a,c)-2l(a,d)\right)+k(b,d)\left(l(b,d)+l(a,c)-2l(c,d)\right)+\\
&k(a,c) \left(l(a,c)+l(b,d)-2l(a,b)\right)+k(a,c)\left(l(a,c)+l(b,d)-2l(a,d)\right)+\\
&k(b,c) \left(l(b,c)+l(a,d)-2l(a,b)\right)+k(b,c)\left(l(b,c)+l(a,d)-2l(b,d)\right)+\\
&k(c,d) \left(l(c,d)+l(a,b)-2l(a,d)\right)+k(c,d)\left(l(c,d)+l(a,b)-2l(b,d)\right)+\\
&k(a,d) \left(l(a,d)+l(b,c)-2l(a,b)\right)+ k(a,d)\left(l(a,d)+l(b,c)-2l(a,c)\right)+\\
&k(b,d) \left(l(b,d)+l(a,c)-2l(a,b)\right)+ k(b,d)\left(l(b,d)+l(a,c)-2l(b,c)\right)+ \\
&k(c,d) \left(l(c,d)+l(a,b)-2l(a,c)\right)+k(c,d)\left(l(c,d)+l(a,b)-2l(b,c)\right).
\end{split}
\end{equation*}
By grouping brackets we obtain 
\begin{equation*}
\begin{split}
&h(z_a,z_b,z_c,z_d) = \\
&k(a,b)\left(2l(a,b)+2l(c,d)-2l(b,c) -2l(b,d)\right) \\
&k(a,c)\left(2l(a,c)+2l(b,d)-2l(b,c) -2l(c,d)\right)+ \\
&k(a,d)\left(2l(a,d)+2l(b,c)-2l(b,d) -2l(c,d)\right)+\\
&k(a,b)\left(2l(a,b)+2l(c,d)-2l(a,c) -2l(a,d)\right)+\\
&k(b,c)\left(2l(b,c)+2l(a,d)-2l(a,c) -2l(c,d)\right)+\\
&k(b,d)\left(2l(b,d)+2l(a,c)-2l(a,d) -2l(c,d)\right)+\\
&k(a,c)\left(2l(a,c)+2l(b,d)-2l(a,b) -2l(a,d)\right)+\\
&k(b,c)\left(2l(b,c)+2l(a,d)-2l(a,b) -2l(b,d)\right)+\\
&k(c,d)\left(2l(c,d)+2l(a,b)-2l(a,d) -2l(b,d)\right)+\\
&k(a,d)\left(2l(a,d)+2l(b,c)-2l(a,b) -2l(a,c)\right)+\\
&k(b,d)\left(2l(b,d)+2l(a,c)-2l(a,b) -2l(b,c)\right)+ \\
&k(c,d)\left(2l(c,d)+2l(a,b)-2l(a,c) -2l(b,c)\right). \\
\end{split}
\end{equation*}
Finally we introduce colours to picture grouping of terms that will cancel each other during integration. 
\begin{equation}
\label{col:brown}
\begin{split}
&h(z_a,z_b,z_c,z_d) = \\
&\big[ k(a,b)\left( {\color{brown} 4l(a,b)}+ {\color{violet} 4l(c,d)}\right)
	 + k(a,c)\left( {\color{brown}4l(a,c)}+ {\color{violet}4l(b,d)}\right)+ \\
&k(a,d)\left( {\color{brown} 4l(a,d)}+{\color{violet} 4l(b,c)} \right)
	+ k(b,c)\left( {\color{ForestGreen} 4l(b,c)}+{\color{Red} 4l(a,d)} \right)+\\
& k(b,d)\left({\color{ForestGreen} 4l(b,d) } +{\color{Red} 4l(a,c)}\right)
	+ k(c,d)\left({\color{ForestGreen} 4l(c,d)}+{\color{Red}4l(a,b)}\right)\big] + \\
\big[ &k(a,b) {\color{brown} \left(- 2l(a,d)-2l(a,c)\right) }
	+ k(a,b){\color{violet} \left(-2l(b,d) -2l(b,c)\right)}+\\
&k(a,c){\color{brown} \left(-2l(a,d) -2l(a,b) \right)}
	+ k(a,c){\color{violet} \left(-2l(c,d) -2l(b,c)\right)}+\\
&k(a,d){\color{brown} \left(-2l(a,c) -2l(a,b)\right)}
	+k(a,d){\color{violet} \left(-2l(c,d) -2l(b,d)\right)}+\\
&k(b,c){\color{Red} \left(-2l(a,c) -2l(a,b) \right)}
	+k(b,c){\color{ForestGreen} \left(-2l(c,d) -2l(b,d)\right)}+\\
&k(b,d){\color{Red} \left(-2l(a,b) -2l(a,d) \right)}
	+ k(b,d){\color{ForestGreen} \left(-2l(b,c) -2l(c,d)\right)}+\\
&k(c,d){\color{Red} \left(-2l(a,d) -2l(a,c)\right)}
	+ k(c,d){\color{ForestGreen} \left(-2l(b,d) -2l(b,c)\right)} \big]\\
\end{split}
\end{equation}
We will show that brown terms of equation \eqref{col:brown} cancel each other. Recall that $h_1(z_1) = \ev  h(z_1,Z_2^*,Z_3^*,Z_4^*)$. Without loss of generality we may assume that we integrate with respect to all variables but $x_a$ and $y_a$. Observe that 
\begin{equation*}
\begin{split}
\ev k(x_a,X_b^*) = \ev k(x_a,X_c^*) = \ev k(x_a,X_d^*) \\
\ev l(y_a,Y_b^*)  = \ev l(y_a,Y_c^*) = \ev l(y_a,Y_d^*) \\
\end{split}
\end{equation*}
Define $q = \ev k(x_a,X_b^*)$, $ p = \ev l(y_a,Y_b^*)$. Therefore, after integration, the brown terms of the equation can be written as
\begin{equation*}
q4p+q4p+q4p+q(-2p-2p) +q(-2p-2p)+q(-2p-2p)=0
\end{equation*}
Similar reasoning shows that red, green and violet terms cancel out.
\end{proof}

\begin{statement}
\label{stm:coreDegeneracy}
 A component of a core function is a canonical core.
\end{statement}
\begin{proof}
 We will use induction by components' index to show that $h_c$ is degenerate. The expected value of the first component is zero, indeed $\ev h_1(Z_1^*)= \ev h(Z_1^*,...,Z_m^*) = 0$. Suppose that for all $c'$ smaller then $c$ degeneracy holds. Using component symmetry it is enough to show that the expected value $\ev h_c(z_1,...,Z_{c}^*)$ is equal to zero. We can write 
\[  \sum_{1 \leq i_1 < ...<i_{c'} \leq c }   h_{c'}(z_{i_1},...,z_{i_{c'}})  
= \sum_{1 \leq i_1 < ...<i_{c'} \leq c-1 }  h_{c'}(z_{i_1},...,z_{i_{c'}}) +  \sum_{1 \leq i_1 < ...<i_{c'-1} < c }  h_{c'}(z_{i_1},...,z_{c}). \]
Now the first sum $\sum_{1 \leq i_1 < ...<i_{c'} \leq c-1 } h_{c'}(z_{i_1},...,z_{i_{c'}})$ does not contain term $z_c$ so integration with respect to $Z_c^*$ does not affect it. On the other hand, by induction assumption $\ev \sum_{1 \leq i_1 < ...<i_{c'-1} < c }  h_{c'}(z_{i_1},...,Z_{c}^*) = 0$.  Obviously $\ev g_c(z_1,...,Z_{c}^*) =  g_{c-1}(z_1,...,z_{c-1})$. Using these observations we obtain 
\begin{equation}
 \begin{split}
  \ev h_c(z_1,...,Z_{c}^*) &=  g_{c-1}(z_1,...,z_{c-1}) -  \sum_{1 \leq i_1 < ...<i_{c-1} \leq c-1 } h_{c-1}(z_{i_1},...,z_{i_{c-1}}) \\
   & - ... -  \sum_{1 \leq i_1 < i_2 \leq c-1 }   h_2(z_{i_1},z_{i_2}) -  \sum_{i=1}^{c-1}  h_1(z_i)  \\
    \end{split}
\end{equation}
Since the  set  $ \{1 \leq i_1 < ...<i_{c-1} \leq c-1 \}$ contains only one sequence,
\begin{equation}
 \begin{split}
   \ev h_c(z_1,...,Z_{c}^*) &= -  h_{c-1}(z_{i_1},...,z_{i_{c-1}}) + [ g_{c-1}(z_1,...,z_{c-1})  \\
   & - ... -  \sum_{1 \leq i_1 < i_2 \leq c-1 }   h_2(z_{i_1},z_{i_2}) -  \sum_{i=1}^{c-1}  h_1(z_i)] =0.\\
 \end{split}
\end{equation}
For this nice simplification we have used definition of the component $h_{c-1}$.
\end{proof}

\begin{lemma}
\label{lem:htilde2}
Under $\mathbf{H_0}$, 
\begin{equation*}
h_2(z_1,z_2) =  \frac 1 6 \tilde{k}(x_1,x_2) \tilde{l}(y_1,y_2)
\end{equation*}
where 
\begin{equation*}
\begin{split}
\tilde{k}(x_1,x_2) &= k(x_1,x_2) - \ev k(x_1,X^*_2) -\ev k(X_1^*,x_2) + \ev k(X_1^*,X_2^*), \\
\tilde{l}(y_1,y_2) &= l(y_1,y_2) - \ev l(y_1,Y^*_2) -\ev l(Y_1^*,y_2) + \ev l(Y_1^*,Y_2^*) 
\end{split}
\end{equation*}
\end{lemma}
\begin{proof}
 We use that $h_2$ is canonial,
 and the exact form of $\ev  h(z_1,z_2,Z_3^*,Z_4^*)$  from \cite{gretton_kernel_2008}, Section A.2, text between equation 12 and 13. 
\end{proof}
 
\begin{corollary}
Under $\mathbf{H_0}$, $h_2 = \frac 1 6 s$.
\end{corollary}

\subsection{Proofs - $U(h_4,Z)$ and $U(h_3,Z)$ converge to zero}
\label{sec:convergenceToZero}
\begin{lemma}
\label{lem:h4null}
If $(Z_t)_{t \in \mathbb{N}+}$ is an absolutely regular process with mixing coefficient decaying faster than $n^{-3}$ ($\beta(n), \theta(n) \leq n^{-3}$), then $n \cdot U(h_4,Z))$ and $n \cdot U(h_3,Z))$ converge to zero in probability.
\end{lemma}

\begin{proof} 
Let  $N:=\{1,\cdots,n\}$,  and let $B$ be a set of all strictly increasing $4$-tuples, $B \subset N^4$. A $U$-statistic can be expressed as sum over elements of $B$,
\[
n \cdot U(h_4,Z))  = \left[ \frac{1}{n^4} {\binom n 4 }^{-1}\right]  \frac{1}{n^3} \sum_{\mathbf b \in B } h_4(Z_b).
\]
If the variance of this random variable goes to zero,
 $$\lim_{n \rightarrow 0}  \ev \left( \frac{1}{n^3} \sum_{\mathbf b \in B } h_4(Z_b) \right)^2 \stackrel{P}{=} 0,$$
then using Chebyshev's inequality we can conclude that it converges to a constant in probability. To show this, we use Lemma 3 from \citet{arcones_law_1998}. We see that the first condition of Theorem 1 from \citet{arcones_law_1998} is met, since $h_4$ is bounded and the mixing coefficient converges to zero. Therefore, by the fact that $h_4$ is canonical, we can use Lemma 3 from \citet{arcones_law_1998}, which states that 
\[
\ev \left( \sum_{\mathbf b \in B } h_4(Z_b) \right)^2 \leq C n^4 M \left(1 + \sum_{m=1}^{n-1} m^3 \beta(m)^{(p-2)/p}\right)
\]
for some $p>2$ and $M = \parallel h \parallel_{\infty}$ . Take $p$ such that $\frac{3 (p-2)}{p}= 2.5$ and use inequality $\beta(m) \leq m^{-3}$ to obtain  
\[
 \sum_{m=1}^{n-1} m^3 \beta(m)^{(p-2)/p} \leq \sum_{m=1}^{n-1} \sqrt m =  O(n^{1.5}).
\]
Therefore 
$$\lim_{n \rightarrow 0}  \ev \left( \frac{1}{n^3} \sum_{\mathbf b \in B } h_4(Z_b) \right)^2  \stackrel{P}{=} \lim_{n \rightarrow 0} \frac{n^{5.5}} {n^6} \stackrel{P}{=} 0.$$ 
We now need to show that $\ev nU(h_4,Z)$ converges to zero. We will use Lemma \ref{lem:divLemma} with $\delta=2$, and that  $\beta(k)^\frac{2}{3} \leq k^{-2}$,  
\begin{equation}
\begin{split}
 \ev nU(h_4,Z) &= \frac n {n(n-1)(n-2)(n-3)}\ev \sum_{1 \leq  a < b < c < d \leq n} h_4(Z_a,Z_b,Z_c,Z_d)  \\ 
 &\leq \frac n {n(n-1)(n-2)(n-3)} \sum_{1 \leq  a < b < c < d \leq n}  M^{\frac{1}{3}} \frac 1 {\max(b-a,c-b,d-c)^2}. 
\end{split}
 \end{equation}
for some constant $M$ as in Lemma \ref{lem:divLemma}. Next
\begin{equation}
\begin{split}
\sum_{1 \leq  a < b < c < d \leq n} & \frac 1 {\max(b-a,c-b,d-c)^2} = \sum_{a=1}^{n-3} \sum_{d=a+3}^{n} \sum_{a< b<c <d} \frac 1 {\max(b-a,c-b,d-c)^2}  \\
& \leq \sum_{a=1}^{n-3} \sum_{d=a+3}^{n} \frac {3^2} {(d-a)^2}   \leq 9 \sum_{a=1}^{n-3} 2 \zeta(2) \leq C n. 
\end{split}
\end{equation}
We have used the fact that $\sum_{d=a+3}^{n} \frac 1 {(d-a)^2}   \leq  2 \zeta(2)$.

The reasoning for $U(h_3,Z)$ is similar.   
\end{proof}

\subsection{Proofs - \citet[Theorem 1]{i._s._borisov_orthogonal_2009} } 
\label{subsub:borisovVolodko}

\begin{Theorem}
\label{borisovVolodko}
Let $m$ be the number of  arguments of a symmetric  kernel $f$. Let one of the following two sets of conditions be fulfilled:
\begin{enumerate}
  \item The stationary sequence $X_i$ satisfies $\theta$-mixing  and
 
  \begin{enumerate}[label*=\arabic*.]
    \item $\sum_{k=1}^{\infty} \phi(k)^{\frac{1}{2}} < \infty$,
    \item $\sup_i \ev | e_i(X_1)|^2 < \infty$.
  \end{enumerate}
  
  \item  The stationary sequence $X_i$ satisfies $\alpha$-mixing. For some $\epsilon > 0$ and for an even number $c \ge 2$ the following holds:
 
  \begin{enumerate}[label*=\arabic*.]
    \item $\sup_i \ev |e_i(X_1)|^{2+\epsilon} \leq \infty$,
    \item $\sum_{k=1}^{\infty} k^{c-2}\alpha^{\epsilon/(c+\epsilon)}(k) < \infty$ 
  \end{enumerate}
\end{enumerate}
where $e_i(X_1)$ are a basis of $L_2(X, F)$.  Then, for any degenerate kernel $f(t_1, . . . , t_m) \in L_2(X_m, F_m)$, under conditions 
\begin{itemize}
\item $\sum_{i_1,...,i_m}^{\infty} |f_{i_1,...,i_m}| < \infty$, where  $f_{i_1,...,i_m}$ are the  coefficient of $f$ in $L_2(X_m, F_m)$,
\item for every collection of pairwise distinct subscripts $(j_1, . . . , j_m)$, the distribution
of $(X_{j_1}, . . . ,X_{j_m})$ is absolutely continuous with respect to the distribution of $(X_1^* , . . . ,X_m^*)$, where $X_i*$ is an independent copy of $X_1$,
\item $e_0=1$ or $\ev  e_i(Z_j) =0$ for all $i$,
\end{itemize}
the following assertion holds:
\begin{equation*}
n^{\frac m 2}U(f,Z) \rightarrow \sum_{i_1,...,i_m}^{\infty} f_{i_1,...,i_m} \prod_{j=1}^{\infty} H_{\nu_j(i_1,...,i_m)}(\tau_j),
\end{equation*}
where  $\tau_j$ is a  centred Gaussian sequence 
with the covariance matrix 
\begin{equation*}
\ev  \tau_k \tau_l = \ev  e_k(X_1) e_l(X_1) + \sum_{j=1}^{\infty} \left[ \ev  e_k(X_1)e_l(X_{j+1}) + \ev  e_l(X_1)e_k(X_{j+1}) \right],
\end{equation*}
$\nu_j(i_1,...,i_m) := \sum_{r=1}^m \delta_{j,i_r}$, and $H_k(x)$ are the Hermite polynomials,
\begin{equation*}
H_k(x) = (-1)^k e^{(x^2/2)} \frac{d^k}{dx^k}(e^{-x^2/2})
\end{equation*}
\end{Theorem}  

\subsection{Proofs - Expected value of the eigenfunctions }
From the eigenvalue equation $\lambda_i \ev e_i(z) = \ev h_2(z,Z_2^*) e_i(Z_2^*)$, $h_2$ degeneracy, and the independence of $Z_1^*$ and $Z_2^*$, we conclude that
\begin{equation*}
\begin{split}
\ev& e_i(Z_1^*) = \frac 1 \lambda_i  \ev h_2(Z_1^*,Z_2^*) e_i(Z_2^*) = \frac 1 \lambda_i \ev [e_i(Z_2^*) \ev (h_2(Z_1^*,Z_2^*)|Z_2^*=z_2)] = \frac 1 \lambda_i \ev [e_i(Z_2^*) \cdot 0] = 0.
\end{split}
\end{equation*}


\end{document}